\documentclass[11pt]{article}

\usepackage{acl}
\usepackage{times}
\usepackage{latexsym}
\usepackage{amsmath,amssymb,mathtools}
\usepackage{booktabs}
\usepackage{tikz}
\usepackage{pgfplots}
\pgfplotsset{compat=1.18}

\usepackage[T1]{fontenc}

\usepackage[utf8]{inputenc}

\usepackage{microtype}
\usepackage{verbatim}
\usepackage{subcaption}

\usepackage{inconsolata}

\usepackage{graphicx}
\usepackage{xurl}

\title{Beyond Behavioral Accuracy: Evaluating Individual-Level Fidelity in LLM Investor Simulation}

\author{
  Jiajie He\textsuperscript{1} \quad
  Yuanhong Jiang\textsuperscript{1} \quad
  Xintong Chen\textsuperscript{3} \quad
  Dongling Ni\textsuperscript{1} \quad
  Wenjin Liu\textsuperscript{2} \quad \\[0.5em]
  \textsuperscript{1}Hithink Research \quad
  \textsuperscript{2}Nanyang Technological University \quad
  \textsuperscript{3}McMaster University \\ [0.5em]
  \texttt{jiajiehe@myhexin.com, yuanhongjiang@myhexin.com, chen54xt@mcmaster.ca} \\
  \texttt{donglingni@myhexin.com,wenjinliu23@outlook.com}
}

\begin{document}

\maketitle 

\begin{abstract}
Large language models (LLMs) are increasingly used as user simulators, but can they faithfully reproduce an individual’s behavior and decision logic over time? We investigate this question in a longitudinal paper-trading study with 80 participants. Using rolling next-day predictions, we evaluate trading activity, buy–sell structure, asset selection, and portfolio outcomes, complemented by controlled input ablations and participant-rated explanation alignment. No evaluated LLM reliably outperforms a simple recent-activity persistence baseline for predicting trade occurrence, while buy–sell structure and asset selection remain difficult to reproduce. Similar activity-level performance can also yield substantially different portfolio trajectories. Ablations show that activity predictions depend primarily on recent trades, whereas asset choices are sensitive to multiple evidence sources. Participants rate explanations as more aligned with their recurring strategies than with their decision-specific rationales. These findings reveal a gap between coarse behavioral agreement and individual-level fidelity, motivating separate evaluation of behavioral accuracy, explanation alignment, and evidence dependence in personalized user simulation. 

\end{abstract}

\section{Introduction}

Large language models (LLMs) are increasingly used to simulate human
behavior in dialogue \cite{chan2023chateval,yoon-etal-2024-evaluating},
recommendation \cite{chen2026realworldhumanbehaviorsimulation},
and interactive decision-making
\cite{10.1145/3586183.3606763,ren-etal-2024-bases}.
User simulators support the evaluation of interactive systems across
diverse user behaviors, but personalized evaluation requires fidelity
to the individual being represented. Plausible actions and aggregate
behavioral similarity provide only partial evidence: a simulator may
match a user's overall activity while acting on different days,
selecting different objects, or generating unsupported explanations.
We therefore ask whether LLMs faithfully reproduce individual behavior
and decision logic over time.

This question is especially important in longitudinal decision-making,
where actions depend on prior behavior, the current environment,
and the consequences of earlier decisions. Financial decision-making
provides an observable setting for examining these dependencies through
trades, holdings, information-seeking activity, and market conditions.
Prior work links investor attention to subsequent stock purchases
\cite{10.1093/rfs/hhm079}, motivating the inclusion of platform
interactions alongside transaction histories.
A simulator may correctly predict whether an investor trades without
recovering the assets or rationale involved. Conversely, it may
accurately characterize a recurring strategy while missing a specific
portfolio adjustment. Behavioral prediction accuracy and alignment
with investor-specific decision patterns therefore require
separate evaluation.

Existing financial-agent research has primarily examined trading
performance, recommendation quality, and aggregate market behavior
\cite{repec:arx:papers:2407.18957,NEURIPS2025_5bf234ec,
10.1145/3805712.3808658}.
More recent work introduces investor simulators with evolving internal
states to evaluate investment advisors over extended interactions
\cite{gong2026shijianbenchdialoguedecisionlonghorizon}.
However, these evaluations do not establish whether a simulator
reproduces the longitudinal decisions of the particular investor
it represents. We address this gap through two complementary questions:
\textbf{how faithfully do LLMs reproduce individual trading behavior,
and what evidence supports their recovery of investor-specific
decision patterns?}

We conduct a longitudinal study with 80 participants in a controlled
paper-trading environment using non-redeemable virtual funds under
real-time market conditions.
Under a rolling next-day prediction protocol, each simulator receives
only information available at the market-close cutoff.
We evaluate trade occurrence, buy--sell structure, asset selection,
and downstream portfolio consequences to distinguish coarse
behavioral agreement from temporally aligned individual fidelity.

To examine investment logic beyond observable actions, we combine
participant-rated explanation alignment, controlled input ablations,
and observational analysis of pre-trade research.
Each participant rates one explanation for their own prediction point,
separately assessing alignment with their recurring investment strategy
and their rationale for the specific decision.
These self-reported judgments measure perceived alignment rather than
faithfulness to the model's internal prediction process
\cite{jacovi-goldberg-2020-towards,madsen-etal-2024-self}.
Ablations measure prediction sensitivity to different evidence sources,
while the observational analysis examines research--trading
associations and the limits of causal interpretation.

No evaluated LLM reliably outperforms a recent-activity persistence
baseline for trade occurrence, while buy--sell structure and asset
selection remain difficult to reproduce and similar activity-level
performance can yield different portfolio trajectories.
Participants report stronger explanation alignment with recurring
strategies than with decision-specific rationales.
Ablations show that activity predictions depend primarily on recent
trades, whereas asset choices are sensitive to multiple evidence sources.
Ticker-specific research predicts subsequent same-ticker trading,
but balance diagnostics and placebo tests do not support a causal
interpretation.

\paragraph{Our Contributions.}
\begin{itemize}

\item To our knowledge, we are the first to systematically evaluate
longitudinal, individual-level fidelity in financial user simulation.
Our framework assesses whether a simulator reproduces the temporally
aligned decisions and portfolio trajectory of the specific investor
it represents.

\item We reveal a hierarchy of behavioral fidelity: coarse agreement
on trading activity does not ensure accurate buy--sell structure,
asset selection, or portfolio outcomes. This demonstrates how
activity-level accuracy can overstate individual-level fidelity.

\item We provide a multi-view analysis of investor-specific decision
patterns by combining participant-rated explanations, controlled
evidence ablations, and observational analysis of pre-trade research.
This distinguishes prediction correctness, perceived explanation
alignment, model evidence dependence, and the limits of causal
interpretation.

\end{itemize}

\section{Task Formulation}
\label{sec:task_formulation}

We formulate personalized investor simulation as a rolling next-day
prediction task: predict a participant's trading activity, buy--sell
structure, ranked assets, and a natural-language explanation using
information available at the prediction cutoff.

\paragraph{Point-in-time information.}
Let $i$ index participants and $t$ index U.S.\ equity trading days.
The cutoff $c_t$ is the market close on day $t$, and the target is
the next trading day $t+1$. The available information is
\begin{equation}
\label{eq:inputs}
\mathcal{I}_{i,t}
=
\left(
\mathbf{u}_{i,t},
\mathbf{r}_{i,t-6:t},
\mathbf{p}_{i,t},
\mathbf{m}_{t}
\right),
\end{equation}
where $\mathbf{u}_{i,t}$ is the historical profile;
$\mathbf{r}_{i,t-6:t}$ contains interactions and executed trades
from the most recent seven trading days;
$\mathbf{p}_{i,t}$ is the portfolio state; and
$\mathbf{m}_{t}$ is the market context.
Longer-term behavioral patterns enter through the profile,
whose construction is described in Appendix~\ref{sec:user_profile}.
All inputs use only information available at or before $c_t$.

\paragraph{Prediction targets.}
The simulator produces
\begin{equation}
\label{eq:prediction}
\begin{aligned}
f_{\theta}(\mathcal{I}_{i,t})
=
\bigl(
&\hat{y}^{\mathrm{trade}}_{i,t+1},
 \hat{y}^{\mathrm{side}}_{i,t+1},\\
&\hat{\mathcal{S}}^{\mathrm{buy}}_{i,t+1},
 \hat{\mathcal{S}}^{\mathrm{sell}}_{i,t+1},
 e_{i,t+1}
\bigr),
\end{aligned}
\end{equation}
where $\hat{y}^{\mathrm{trade}}_{i,t+1}$ indicates whether trading
is predicted and $e_{i,t+1}$ is the explanation.
For a predicted trading day, $\hat{y}^{\mathrm{side}}_{i,t+1}$
is \textsc{buy-only}, \textsc{sell-only}, or \textsc{buy-and-sell},
with at most five ranked tickers per applicable side.
For a predicted non-trading day, the side is not applicable and
both asset lists are empty.

A point-in-time candidate set $\mathcal{C}_{i,t}$ is constructed
from current holdings, recent trades, and recently viewed assets.
The model receives this set but may name assets outside it when
supported by pre-cutoff evidence; candidate filtering is applied
during the main asset evaluation.

\begin{figure*}[t]
  \centering
  \includegraphics[width=\textwidth]{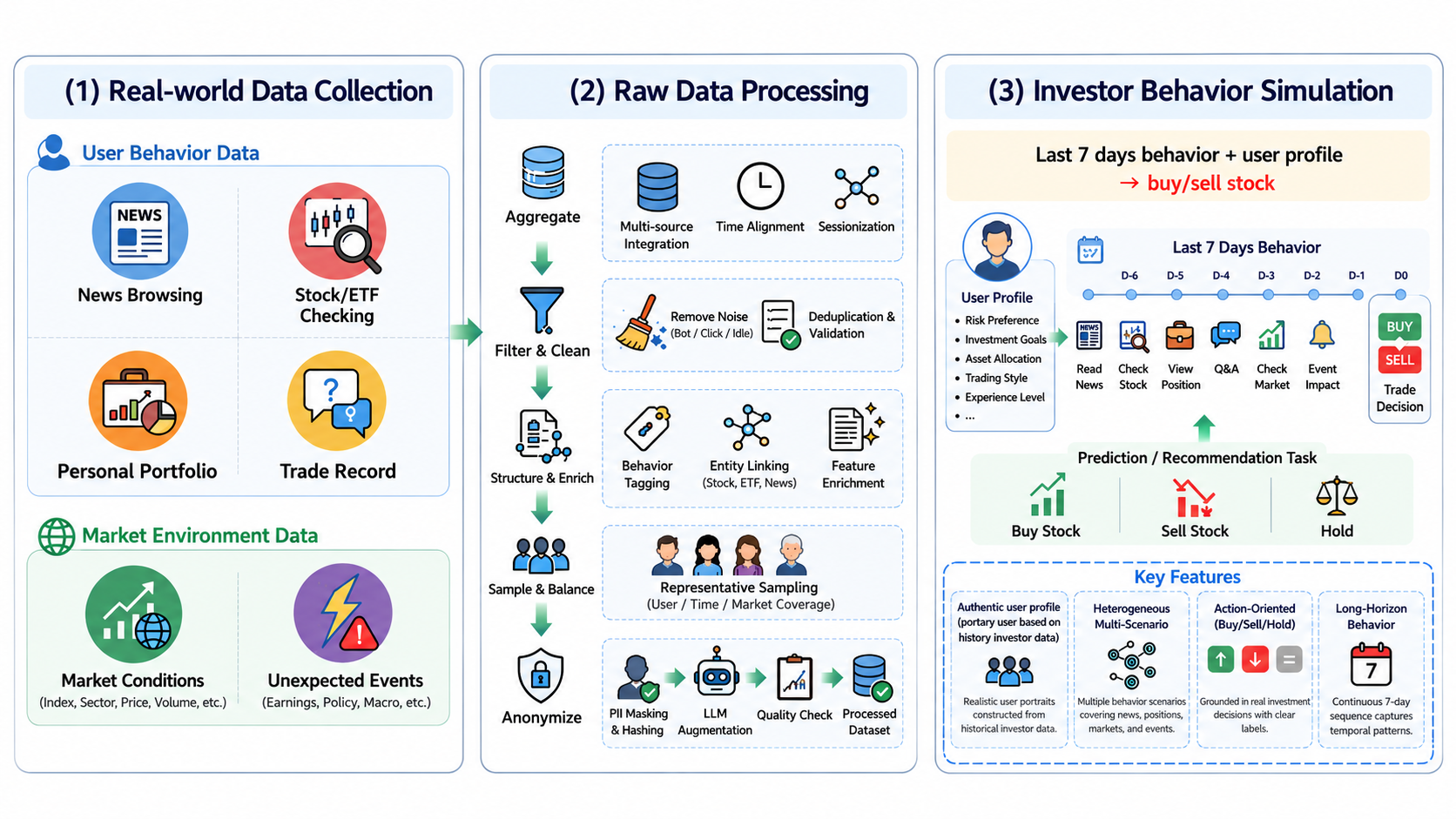}
  \caption{
  Rolling investor simulation. Evidence available at market close
  on day $t$ informs predictions for the next trading day $t+1$;
  newly observed records update subsequent inputs.
  }
  \label{fig:pipeline_user_prediction}
\end{figure*}

\paragraph{Rolling protocol.}
After each target day, newly observed records update the inputs
for the next prediction, as illustrated in
Figure~\ref{fig:pipeline_user_prediction}.

\section{Task Evaluation}
\label{sec:task_Evaluation}

We evaluate behavioral fidelity, participant-rated explanation
alignment, and evidence dependence, and examine the association
between pre-trade research and subsequent trading.

\subsection{Behavioral Fidelity}
\label{sec:behavioral_fidelity}

\paragraph{Trade occurrence.}
For an observed participant--day,
$y^{\mathrm{trade}}_{i,t+1}=\mathbb{I}[K_{i,t+1}>0]$,
where $K_{i,t+1}$ is the number of executed trades.
We report accuracy, macro-F1, and class-specific precision and recall.

\paragraph{Buy--sell structure.}
On all observed trading days, $y^{\mathrm{side}}_{i,t+1}$ is
\textsc{buy-only}, \textsc{sell-only}, or \textsc{buy-and-sell}.
We report accuracy and macro-F1,
averaging F1 over these three classes.
Evaluation includes missed trades: a no-trade prediction on an
observed trading day counts as a side error.

\paragraph{Asset selection.}
Buy and sell rankings are evaluated separately on all observed days
containing executions on the corresponding side, regardless of
trade-occurrence or side-prediction correctness.
A \textsc{buy-and-sell} day contributes to both evaluations.
Predicted lists are filtered to $\mathcal{C}_{i,t}$, deduplicated
while preserving order, and limited to five assets; the observed
asset set remains unfiltered.

Hit@5\cite{10.1145/3038912.3052569}
indicates whether a predicted list contains at least one observed
asset, and NDCG@5\cite{10.1145/582415.582418} measures ranking
quality with binary relevance.
A missing or empty list receives zero for both metrics.
Scores are averaged over eligible participant--days for each side.
Candidate coverage is the proportion of eligible days with at least
one observed asset in $\mathcal{C}_{i,t}$.

\paragraph{Portfolio consequences.}
We compare observed and simulated portfolio paths and terminal
returns using common adjusted closing prices.
The simulated portfolio starts from the same initial holdings and
cash, and evolves through predicted buys and sells, with transaction
sizes calibrated from observed portfolio changes over the available
period.
These results are exploratory retrospective comparisons rather
than cutoff-safe portfolio forecasts.

\subsection{Participant-Rated Explanation Alignment}
\label{sec:explanation_alignment}

Each of the 80 participants rates one model-generated explanation
for their own prediction point on two five-point ordinal scales:
\textbf{strategy alignment}, measuring consistency with their
recurring investment strategy, and
\textbf{decision-rationale alignment}, measuring consistency with
their rationale for the specific decision.
Higher ratings indicate stronger perceived alignment.
We analyze these self-reported ratings separately from prediction
correctness.

\subsection{Evidence Dependence and Behavioral Association}
\label{sec:evidence_dependence}

\paragraph{Input ablations.}
For each evidence group $g$, we compare predictions from
$\mathcal{I}_{i,t}$ and the corresponding ablated input
$\mathcal{I}_{i,t}^{(-g)}$ on matched participant--date pairs.
We separately withhold recent trades, portfolio information,
recent platform interactions, market context, and the long-horizon
profile.
We report activity-accuracy changes and activity flip rates on
all scored days, side-accuracy changes and side/list flip rates
on observed trading days, and buy/sell Hit@5 changes on the
corresponding observed-side evaluation sets.
An asset-list flip occurs when either ordered buy or sell list
differs between conditions.
Unlike the main candidate-filtered evaluation, ablation asset
metrics use the model-named lists before candidate filtering.
These comparisons measure the effects of evidence removal.

\paragraph{Auxiliary position-action evaluation.}
We additionally evaluate a day-level primary position-action label
$y^{\mathrm{pos}}_{i,t+1}$ with five categories:
\textsc{open}, \textsc{add}, \textsc{reduce}, \textsc{close},
and \textsc{unchanged}.
We report accuracy on observed position labels under full and
ablated inputs, with label construction detailed in the Appendix.

\paragraph{Observed research and trading.}
Using a target-trial emulation framework \cite{hernan2016using},
we examine the association between ticker-specific research and
next-trading-day same-ticker execution.
We report the adjusted trading-probability difference alongside
covariate-balance, overlap, and placebo diagnostics to assess
the limits of causal interpretation.

\section{Experiments}
\label{sec:experiments}

We conduct a pilot evaluation of individual-level fidelity
and failure modes in LLM-based financial user simulation.
Experimental details are provided in
Appendix~\ref{sec:experiment_setting}.

\subsection{RQ1: How Faithful Are LLMs as Financial User Simulators?}
\label{sec:rq1}

We evaluate behavioral fidelity across trade occurrence,
buy--sell structure, asset selection, and portfolio consequences.
This evaluation examines whether agreement on trading activity
extends to temporally aligned, asset-specific decisions and
their downstream outcomes.

\paragraph{Finding 1: Aggregate activity agreement masks
temporal errors.}

Non-trading accounts for 75.22\% of scored participant--days,
making overall accuracy sensitive to conservative no-trade
predictions. GPT-5.6-Sol achieves the highest LLM accuracy
(79.18\%) but only 34.85\% trade recall, whereas Gemini-3-Flash
achieves lower accuracy (68.60\%) but substantially higher
trade recall (76.55\%). A recent-activity persistence baseline
achieves 79.82\% accuracy and 71.45\% macro-F1, exceeding all
evaluated LLM point estimates on both metrics. Participant-level
bootstrap comparisons do not establish a reliable LLM
improvement over this baseline
(Table~\ref{tab:trade_side}).

\begin{figure*}[t]
\centering
\includegraphics[width=.96\textwidth]{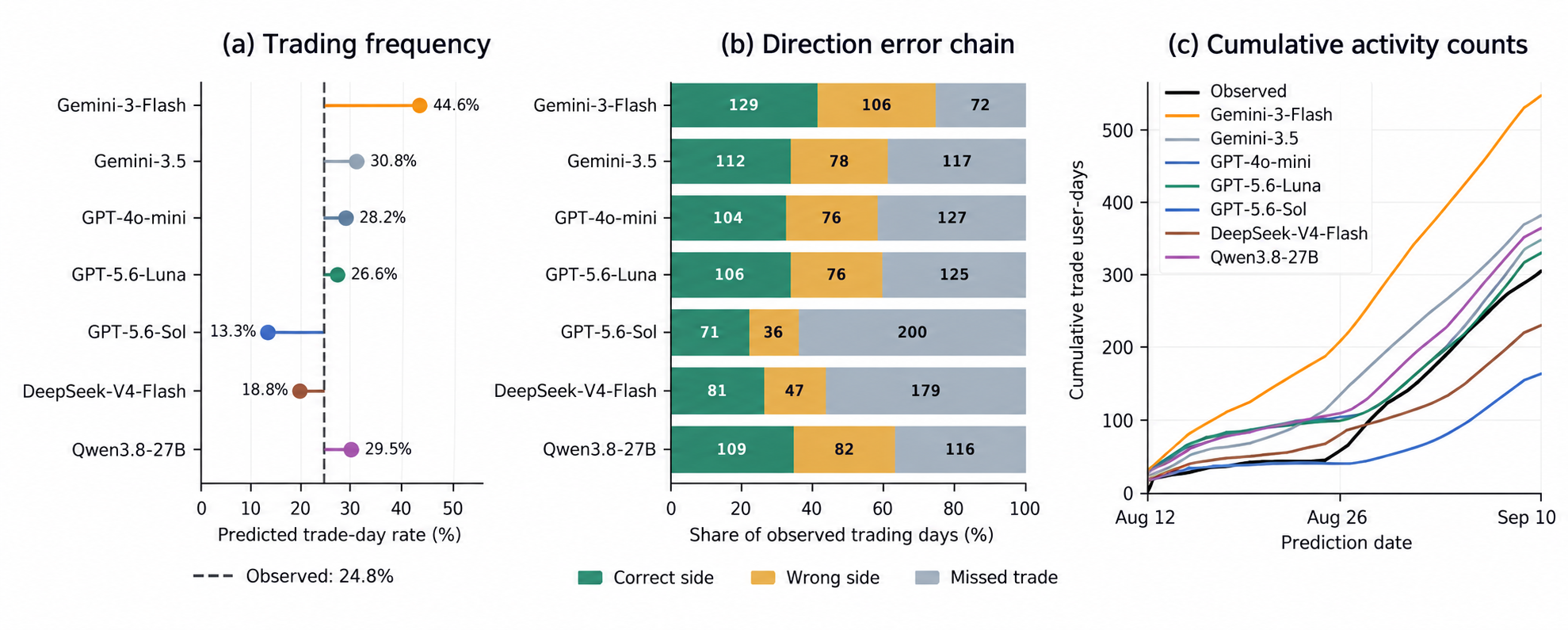}
\caption{
Activity and action-structure mismatch.
(a) Predicted versus observed trading frequency.
(b) Observed trading days partitioned into correct side
predictions, incorrect side predictions, and missed trades.
(c) Cumulative observed and predicted trading activity.
}
\label{fig:structural-bias}
\end{figure*}

Matching aggregate trading frequency also fails to establish
temporal fidelity. GPT-5.6-Luna predicts trading on 26.6\% of
participant--days, close to the observed 24.8\%, yet achieves
only 59.28\% trade recall. Thus, similar activity rates can
coexist with substantial disagreement about when individual
participants trade
(Figure~\ref{fig:structural-bias}(a,c)).

\begin{table*}[t]
\centering
\footnotesize
\caption{
Trade-occurrence and end-to-end buy--sell structure prediction
(\%). Activity metrics use 1,239 scored participant--days;
side metrics use all 307 observed trading days, counting
no-trade predictions as incorrect.
Bold indicates the highest point estimate per metric.
}
\label{tab:trade_side}

\renewcommand{\arraystretch}{0.95}

\begin{tabular*}{\textwidth}{
@{\extracolsep{\fill}}
lrrrrr
@{}
}
\toprule
& \multicolumn{3}{c}{Trade Occurrence}
& \multicolumn{2}{c}{Side Structure} \\
\cmidrule(lr){2-4}
\cmidrule(lr){5-6}
Model
& Acc.
& Macro-F1
& \shortstack{Trade\\Recall}
& Acc.
& Macro-F1 \\
\midrule

Gemini-3-Flash
& 68.60 & 65.34 & \textbf{76.55}
& \textbf{42.02} & \textbf{37.88} \\

Gemini-3.5-Flash
& 75.06 & 68.94 & 61.89
& 36.48 & 34.05 \\

GPT-4o-mini
& 76.11 & 69.32 & 58.63
& 33.88 & 35.80 \\

GPT-5.6-Luna
& 78.05 & 71.23 & 59.28
& 34.53 & 34.53 \\

GPT-5.6-Sol
& 79.18 & 66.24 & 34.85
& 23.13 & 28.46 \\

DeepSeek-V4-Flash
& 77.08 & 66.38 & 41.69
& 26.38 & 29.39 \\

Qwen3.8-27B
& 76.59 & 70.39 & 62.21
& 35.50 & 35.42 \\

\midrule

Always no-trade
& 75.22 & 42.93 & 0.00
& -- & -- \\

Recent Activity Persistence
& \textbf{79.82} & \textbf{71.45} & 51.79
& -- & -- \\

Always buy-and-sell
& -- & -- & --
& 41.37 & 19.51 \\

\bottomrule
\end{tabular*}
\end{table*}

\paragraph{Finding 2: Trade detection does not ensure
action or asset fidelity.}

Buy-and-sell days account for 41.37\% of observed trading
days. The best model achieves 42.02\% side accuracy,
close to the 41.37\% majority baseline, although its
macro-F1 improves from 19.51\% to 37.88\%
(Table~\ref{tab:trade_side}).
Sell-only decisions remain particularly difficult,
with end-to-end recall of just 4.23--8.45\%.
Errors arise both from missed trades and from incorrect
buy--sell structures on detected trading days
(Figure~\ref{fig:structural-bias}(b)).

Asset rankings are evaluated on all observed buying
and selling days, regardless of activity or side-prediction
correctness. The best Hit@5 reaches 40.7\% for buys
and 42.9\% for sells
(Table~\ref{tab:ticker_retrieval}).
Candidate coverage further constrains retrieval:
at least one observed asset appears in the candidate set
on 73.31\% of buying days and 91.92\% of selling days.
The gap between coverage and retrieval,
particularly for sells, shows that candidate availability
alone does not ensure accurate asset selection.

\begin{table}[t]
\centering
\footnotesize
\caption{
Candidate-filtered asset retrieval on all observed trading days, without conditioning
on prediction correctness.
Hit@5 is a percentage; NDCG@5 uses a 0--1 scale.
Bold indicates the best result per column.
}
\label{tab:ticker_retrieval}

\setlength{\tabcolsep}{2.2pt}
\renewcommand{\arraystretch}{0.92}

\begin{tabular}{@{}lrrrr@{}}
\toprule

Model &
\shortstack{Buy\\Hit@5} &
\shortstack{Buy\\NDCG@5} &
\shortstack{Sell\\Hit@5} &
\shortstack{Sell\\NDCG@5} \\

\midrule

Gemini-3-Flash
& \textbf{40.7} & 0.259
& \textbf{42.9} & \textbf{0.272} \\

Gemini-3.5-Flash
& 39.0 & \textbf{0.260}
& \textbf{42.9} & 0.265 \\

GPT-4o-mini
& 37.7 & 0.247
& 27.8 & 0.164 \\

GPT-5.6-Luna
& 35.2 & 0.223
& \textbf{42.9} & 0.257 \\

GPT-5.6-Sol
& 22.9 & 0.150
& 32.3 & 0.219 \\

DeepSeek-V4-Flash
& 25.8 & 0.164
& 35.4 & 0.227 \\

Qwen3.8-27B
& 39.0 & 0.250
& 39.9 & 0.232 \\

\bottomrule
\end{tabular}
\end{table}

\paragraph{Finding 3: Similar activity fidelity can
coexist with divergent portfolio trajectories.}

\begin{figure*}[t]
\centering
\includegraphics[
  width=\textwidth
]{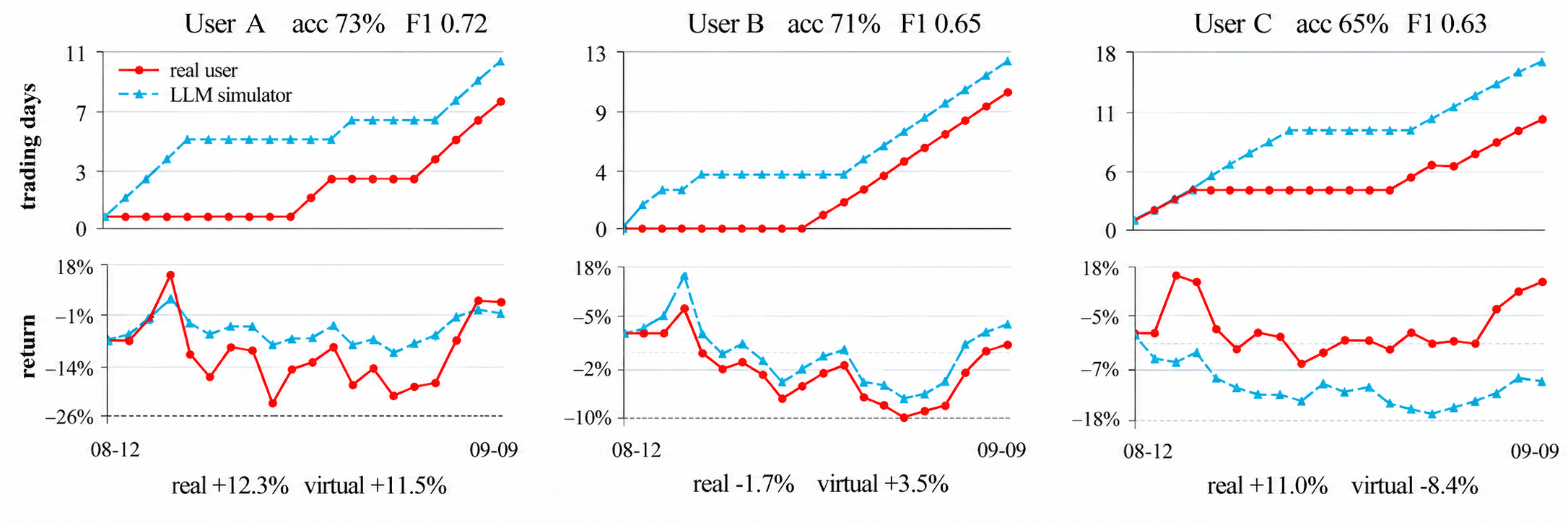}
\caption{Illustrative comparisons of observed and simulated
trading activity and portfolio trajectories.
Similar activity-level fidelity can yield different portfolio outcomes.}
\label{fig:rq1-return-curves}
\end{figure*}

Figure~\ref{fig:rq1-return-curves} compares participants
B and C, whose activity macro-F1 scores are similar
(0.65 and 0.63).
For participant B, observed and simulated terminal returns
differ by 1.8 percentage points (1.7\% versus 3.5\%);
for participant C, the difference reaches 19.4 points
(+11.0\% versus -8.4\%).

These comparisons use transaction sizes calibrated
retrospectively over the available observation period.
They illustrate how similar activity-level fidelity
can coexist with different portfolio outcomes,
rather than constituting independent point-in-time
return forecasts or establishing the prevalence
of divergence across participants.

\subsection{RQ2: Does the Simulator Recover User Investment Logic?}
\label{sec:rq2}

We examine investor-specific decision patterns through
input sensitivity and participant-rated explanation
alignment. These analyses assess evidence dependence
and perceived alignment, rather than directly identifying
latent decision mechanisms. An observational research--trading
analysis provides additional context about participants'
pre-trade behavior.

\paragraph{Finding 4: Activity predictions depend primarily
on recent trades, while asset outputs show broader
evidence sensitivity.}

We conduct paired input ablations with GPT-5.6-Sol,
separately withholding recent trades, portfolio information,
recent platform interactions, market context, and the
long-horizon profile.
Removing recent trades reduces trade-occurrence accuracy
by 23.49 percentage points
(95\% CI: $[-33.00,-13.65]$), whereas removing any other
individual evidence source changes accuracy by at most
1.40 points (Table~\ref{tab:rq2-ablation}).

This pronounced dependence suggests that the simulator
draws heavily on recent trading regularities when predicting
whether an investor will trade. Such regularities can
provide useful predictive signals, but accurate activity
predictions alone do not establish recovery of the
investor-specific rationale behind each decision.

Asset outputs exhibit broader evidence sensitivity.
Withholding any evidence group changes more than half
of the predicted ticker lists. Removing the long-horizon
profile, for example, changes 60.4\% of ticker lists
while altering only 6.7\% of trade-occurrence predictions
and reducing activity accuracy by just 0.23 percentage
points. This contrast indicates that predictions about
whether to trade can remain relatively stable even as
the proposed assets change substantially.

These findings suggest that coarse activity prediction
and asset selection exhibit different evidence
dependencies. However, ticker-list changes do not
necessarily imply reduced asset accuracy or successful
integration of investment logic, and some changes may
arise from altered buy--sell predictions. The broader
sensitivity of asset outputs motivates examining how
user simulators integrate heterogeneous investor evidence
when predicting specific investment decisions.

\begin{table}[t]
\centering
\footnotesize
\caption{
Evidence ablation with GPT-5.6-Sol.
Accuracy changes are relative to the paired full-input
condition. Activity flips indicate changes in trade-occurrence
predictions; ticker flips indicate changes in either
ordered buy or sell list.
}
\label{tab:rq2-ablation}

\renewcommand{\arraystretch}{0.88}

\begin{tabular*}{\columnwidth}{
@{\extracolsep{\fill}}
lrrr
@{}
}
\toprule

Withheld evidence &
\shortstack{$\Delta$ Acc.\\(pp)} &
\shortstack{Activity\\flip} &
\shortstack{Ticker\\flip} \\

\midrule

Recent trades
& \textbf{-23.49}
& \textbf{36.5\%}
& \textbf{76.6\%} \\

Portfolio
& -1.40 & 7.0\% & 58.6\% \\

7-day behavior
& -0.23 & 4.4\% & 56.8\% \\

Market context
& -1.16 & 4.0\% & 57.7\% \\

Long-term profile
& -0.23 & 6.7\% & 60.4\% \\

\bottomrule
\end{tabular*}
\end{table}

\paragraph{Finding 5: Strategy alignment exceeds
decision-rationale alignment.}

Each of the 80 participants rated one GPT-5.6-Sol
explanation generated for their own prediction point,
separately assessing alignment with their recurring
investment strategy and their rationale for the
specific decision.

Participants report stronger alignment with recurring
strategies than with decision-specific rationales
(Table~\ref{tab:explanation_alignment}).
Mean ratings are 3.40 and 2.43, respectively;
ratings of 4--5 account for 56.25\% of strategy
evaluations but only 15.00\% of decision-rationale
evaluations.

These results indicate that participants more frequently
perceive generated explanations as consistent with broad
investment patterns than with the rationale for a specific
decision. This gap complements the evidence-ablation
findings: capturing recurring behavioral patterns does
not necessarily translate into explanations aligned
with individual investment decisions. Representative qualitative cases are provided in
Appendix~\ref{app:explanation_alignment_cases}.

\begin{table}[t]
\centering
\footnotesize
\caption{
Participant-rated explanation alignment
($n=80$; one explanation per participant).
Ratings range from 1 to 5; higher indicates stronger
perceived alignment.
}
\label{tab:explanation_alignment}

\renewcommand{\arraystretch}{1.05}
\setlength{\tabcolsep}{3pt}

\begin{tabular*}{\columnwidth}{
@{\extracolsep{\fill}}
lrrrrrr
@{}
}
\toprule

Dimension & 1 & 2 & 3 & 4 & 5 & Mean \\

\midrule

\shortstack[l]{Strategy\\alignment}
& 8 & 3 & 24 & 39 & 6 & 3.40 \\

\shortstack[l]{Decision-rationale\\alignment}
& 18 & 22 & 28 & 12 & 0 & 2.43 \\

\bottomrule
\end{tabular*}
\end{table}

\paragraph{Finding 6: Research escalation is associated
with subsequent trading, but causation is not established.}

Across 15,125 participant--ticker--date opportunities,
including 395 research escalations, intensified research
is associated with an adjusted 3.94 percentage-point
increase in next-day same-ticker trading probability
(95\% CI: $[1.92,6.98]$).
However, post-weighting covariate balance remains inadequate
(maximum weighted SMD $=0.324$, above the pre-specified
0.10 threshold), and the future-behavior placebo remains
positive (+5.12 pp; 95\% CI: $[1.58,10.07]$).
Research escalation may signal developing investment intent,
but these diagnostics do not support interpreting it as
the cause of subsequent trading
(Figure~\ref{fig:rq2-target-trial}).

\begin{figure}[t]
\centering
\resizebox{\columnwidth}{!}{%
\begin{tikzpicture}
\begin{axis}[
    width=0.62\columnwidth,
    height=3.5cm,
    scale only axis,
    xmin=-1.5,
    xmax=10.8,
    ymin=0.65,
    ymax=6.35,
    xlabel={Probability difference (pp)},
    xlabel style={font=\scriptsize, yshift=3pt},
    xtick={0,2.5,5,7.5,10},
    ytick={1,2,3,4,5,6},
    yticklabels={
        {Future placebo},
        {Wrong-ticker control},
        {Pre-trend placebo},
        {Sell},
        {Buy},
        {Same-ticker trade}
    },
    tick label style={font=\scriptsize},
    yticklabel style={
        font=\scriptsize,
        align=right
    },
    xmajorgrids=true,
    grid style={gray!20},
    axis x line=bottom,
    axis y line=left,
    clip=true
]

\addplot[dashed, thick, black]
coordinates {(0,0.65) (0,6.35)};

\addplot[thick, blue]
coordinates {(1.92,6) (6.98,6)};
\addplot[only marks, mark=*, blue]
coordinates {(3.94,6)};

\addplot[thick, blue]
coordinates {(0.87,5) (5.81,5)};
\addplot[only marks, mark=*, blue]
coordinates {(2.95,5)};

\addplot[thick, blue]
coordinates {(1.68,4) (6.94,4)};
\addplot[only marks, mark=*, blue]
coordinates {(3.86,4)};

\addplot[thick, gray]
coordinates {(-0.87,3) (2.07,3)};
\addplot[only marks, mark=*, gray]
coordinates {(0.34,3)};

\addplot[thick, gray]
coordinates {(-0.40,2) (3.60,2)};
\addplot[only marks, mark=*, gray]
coordinates {(1.41,2)};

\addplot[thick, red]
coordinates {(1.58,1) (10.07,1)};
\addplot[only marks, mark=*, red]
coordinates {(5.12,1)};

\end{axis}
\end{tikzpicture}%
}
\caption{Research escalation predicts subsequent trading,
but placebo tests limit causal interpretation.}
\label{fig:rq2-target-trial}
\end{figure}

Together, these findings reveal a gap between capturing
behavioral regularities and faithfully representing
investor-specific decision patterns. Trade-occurrence
predictions depend strongly on recent trading history,
while asset outputs exhibit sensitivity to multiple
sources of investor evidence. Participants also perceive
stronger explanation alignment with recurring strategies
than with decision-specific rationales.

These observations highlight the importance of evaluating
investment logic beyond behavioral prediction accuracy.
Faithful investor simulation requires attention not only
to when investors trade, but also to how their historical
behavior, portfolio context, preferences, and market
information relate to specific investment decisions.
Our findings motivate such multi-view evaluation without
claiming to identify participants' latent decision mechanisms.

More experimental details are shown in Appendix~\ref{app:evaluation_details}.

\section{Related Work}

\subsection{LLMs as Human Simulators}

LLMs have been used to simulate behavior in
dialogue~\cite{chan2023chateval},
recommendation~\cite{chen2026realworldhumanbehaviorsimulation,10.1145/3708985,10.1145/3589334.3645537},
and autonomous driving~\cite{jin2024surrealdriverdesigningllmpoweredgenerative}.
Generative Agents~\cite{10.1145/3586183.3606763}
and BASES~\cite{ren-etal-2024-bases} extend user simulation
to social behavior and web search.
For personalized simulation, behavioral plausibility must be
distinguished from fidelity to a particular individual:
matching aggregate patterns does not establish agreement
with that user's decisions over time.
Our study examines this distinction through longitudinal
trading behavior, evidence dependence, and participant-rated
explanation alignment.

\subsection{Financial Agents and Investor Simulation}

Early financial user simulators supported personalized
dialogue management for financial product
recommendation~\cite{10.1145/3350546.3352501}.
StockAgent~\cite{repec:arx:papers:2407.18957} simulates
investor trading under dynamic market conditions,
while TwinMarket~\cite{NEURIPS2025_5bf234ec} examines
how individual trading decisions generate collective
market behavior.
Conv-FinRe~\cite{10.1145/3805712.3808658} constructs
longitudinal financial recommendation scenarios from
market data and human decision trajectories.
ShiJianBench~\cite{gong2026shijianbenchdialoguedecisionlonghorizon}
introduces investor simulators with evolving internal
states to evaluate investment advisors over extended
interactions.

These studies address financial-agent performance,
recommendation, and simulated investor behavior.
Our focus is whether a simulator reproduces the temporally
aligned decisions of the specific investor it represents.
We evaluate this individual-level fidelity alongside
participant-perceived explanation alignment and prediction
sensitivity to investor-specific evidence.

\section{Conclusion}

Our study highlights a central challenge in personalized
investor simulation: reproducing recurring behavior
does not establish understanding of the rationale
behind individual decisions.
Progress therefore requires connecting user-specific
evidence to evolving investment choices and validating
both predicted actions and decision rationales
over time.Future work will explore dynamically updated
user representations and evidence-grounded reasoning that connects historical behavior, portfolio
constraints, and market information to specific investment decisions. Repeated participant feedback and prospective evaluation will help assess whether
these methods improve both decision prediction and rationale alignment as users’ investment strategies evolve.

\section{Limitations}

Our findings are based on 80 participants in a controlled paper-trading environment. The cohort, observation period, and absence of real financial stakes may limit generalization to broader populations, market conditions, and real-world trading. Ablation and explanation analyses cover a single evaluated model due to funding limit and one explanation rated per participant due to high human label fee may produce bias in the final experimental results. Input sensitivity and self-reported alignment do not identify internal reasoning or establish consistent alignment across multiple decisions.

\section{Ethical Considerations}

Our study involves behavioral records from 80 participants
in a controlled paper-trading environment using virtual
funds. Although the study does not involve real financial
transactions, trading histories and platform interactions
may reveal sensitive behavioral patterns. We therefore
de-identify participant records and avoid releasing direct
personal identifiers. However, residual re-identification
risks may remain.

Furthermore, our findings demonstrate limitations in the
individual-level fidelity of LLM-based investor simulation.
Such simulators should not be treated as reliable substitutes
for real investors or used as the sole basis for consequential
financial decisions. Potential misuse for intrusive behavioral
profiling should also be considered.

\bibliography{custom}

\appendix

\appendix

\section{Experimental Setting}
\label{sec:experiment_setting}

\subsection{Models and Behavioral Baselines}

We evaluate seven LLM configurations: GPT-5.6-Sol\cite{openai2026gpt56},
GPT-5.6-Luna\cite{openai2026gpt56},
GPT-4o-mini\cite{openai2024gpt4omini},
Gemini-3-Flash\cite{gemini3flash2025google},
Gemini-3.5-Flash\cite{gemini35flash2026},
DeepSeek-V4-Flash, and Qwen3.8-27B.
All main-model comparisons use the same participant--date instances
and observed target labels.

The recent-activity persistence baseline predicts trading on day
$t+1$ if at least one execution is recorded on trading day $t$,
and otherwise predicts no trading.
The always-no-trade baseline predicts inactivity for every scored day.
For side classification, the always-buy-and-sell baseline predicts
\textsc{buy-and-sell} on every observed trading day.


\subsection{Dataset}

\paragraph{Data source.}
We recruited 80 volunteer participants for a controlled paper-trading
study under real-time market conditions.
Participants could buy, sell, rebalance, or remain inactive using
non-redeemable virtual funds.
The study did not access their real brokerage accounts, credentials,
funds, or transaction records.
We recorded trajectories over a three-month collection period,
including more than 230K interactions across market and news browsing,
stock/ETF viewing, simulated trading, virtual portfolio states,
and platform-chatbot interactions.


\paragraph{Privacy protection.}
We de-identify participant-generated behavioral records used for
analysis and prepared for release.
A locally deployed DeepSeek-V4-Flash model~\cite{deepseekai2026deepseekv4highlyefficientmilliontoken}
assists in detecting identifiers in free-text and account-associated
fields, which are replaced with standardized placeholders such as
\texttt{[NAME]} and \texttt{[AGE]}.
Manual quality assurance follows automated processing.
The processed research data exclude raw personal identifiers and
real financial-account records; virtual portfolio states remain
available for the reported analyses.

\paragraph{Data processing.}
We remove duplicated, incomplete, and noisy records and organize
interactions into chronological participant sequences.
Pre-cutoff behavior supports profile construction, while cleaned
financial-news text is summarized using
\textbf{DeepSeek-V4}\cite{deepseekai2026deepseekv4highlyefficientmilliontoken}
and inserted at the corresponding positions in the sequence.
The cohort includes active and inactive participants with different
trading styles and levels of financial experience.

\paragraph{Evaluation coverage.}
The three-month collection period is distinct from the rolling
prediction evaluation period.
The main experiments contain 1,239 predictions from 80 participants; Table~\ref{tab:appendix-coverage} summarizes the evaluation sets.
Buying and selling sets overlap because a day with both sides
contributes to both asset evaluations.

\begin{table}[t]
\centering
\footnotesize
\caption{Coverage of prediction and evaluation sets.}
\label{tab:appendix-coverage}

\renewcommand{\arraystretch}{0.78}

\begin{tabular*}{\columnwidth}{
@{\extracolsep{\fill}}
lr
@{}
}
\toprule
Quantity & Count \\
\midrule
Prediction instances & 1,239 \\
Trading days & 307 \\
Non-trading days & 932 \\
Buying days & 236 \\
Selling days & 198 \\
Buy-and-sell days & 127 \\
\bottomrule
\end{tabular*}
\end{table}

\subsection{Prediction Protocol and Prompt}

Each prediction targets the next U.S.\ equity trading day after
the market-close cutoff.
Model inputs contain the cutoff-specific profile, recent interactions
and trades, portfolio state, and market context.
Target-day executions and subsequent portfolio records are reserved
for evaluation and are not provided as prediction evidence.

The prediction instructions specify trade occurrence, buy--sell
structure, at most five ranked assets per applicable side,
and an explanation.
A no-trade answer has no applicable side and empty asset lists.
The candidate list is available to the model, but evidence-supported
out-of-list assets may be named.
Candidate-filtered and model-named lists are distinguished in scoring.

%

\section{User Profile Construction}
\label{sec:user_profile}

We construct user profiles using a staged multi-agent
pipeline~\cite{wang2026llmbaseduserpersonasrecommendations,
10.1007/978-3-031-05409-9_40}.
Each profile summarizes observable investment behavior using only
information available at the prediction cutoff. Figure~\ref{fig:profile-example} shows the specific example.

\paragraph{Evidence preparation.}
For participant $i$ at cutoff $c_t$, we collect interaction events,
executed trades, portfolio snapshots, account states, and available
conversation records. Records are aligned by participant and
timestamp, excluding information unavailable at $c_t$.
A deterministic feature module computes activity statistics,
transaction amounts, portfolio concentration, drawdown, and
holding patterns directly from these records.

\paragraph{Agent analysis and synthesis.}
An orchestrator routes the structured evidence to specialized
agents for risk behavior, investment logic, revealed trading
preferences, responses to gains and losses, and investor activity
patterns. Each agent receives a restricted evidence view and
returns conclusions with supporting evidence and uncertainty
indicators. Their outputs are merged and checked for conflicts
before synthesis into an overview, dimension-level summaries,
evidence-grounded highlights, and caveats. Unsupported attributes
remain unknown.

\paragraph{Temporal representation.}
Profiles are stored by participant and cutoff. Subsequent updates
use only newly available evidence, preserving the information
boundary for each prediction. The simulator receives the
corresponding profile, recent interactions and trades, portfolio
state, and market context, as defined in
Equation~\ref{eq:inputs}. The profile summarizes longer-term
patterns, while the recent evidence blocks capture conditions
relevant to the next trading day.

\begin{figure}[t]
\centering
\begingroup
\setlength{\fboxsep}{7pt}
\setlength{\fboxrule}{0.4pt}
\fbox{%
\parbox{\dimexpr\linewidth-2\fboxsep-2\fboxrule\relax}{%
\small
\textbf{Base Profile: Participant $U_{\mathrm H}$}\\

\medskip
\textbf{Overview.}
An active, research-first participant who explores broadly
but allocates selectively. Short-horizon trading coexists
with relatively stable holdings.

\medskip
\textbf{Risk behavior.}
Average invested-capital ratio: 95\%;
observed account maximum drawdown: 10.2\%.
Self-reported risk tolerance is unknown.

\medskip
\textbf{Investment logic.}
Research centers on price checking and technical information.
Preceding research is observed for 88.5\% of trades, with a
median research-to-trade interval of 83.3 hours.
Average holding duration is 5.43 days.

\medskip
\textbf{App behavior.}
Research covers 131 securities.
Price checking accounts for 83.1\% of research interactions;
the research-to-holding conversion ratio is 4.1\%.

\medskip
\textbf{Responses to gains and losses.}
Losing positions are held 1.21 times as long as winning
positions, on average. Averaging down is observed;
momentum chasing and panic selling remain unknown.

\medskip
\textbf{Assessment and caveats.}
The profile describes an active opportunity explorer.
Self-reported objectives, experience, investment horizon,
and life stage are unavailable. Account drawdown may
reflect deposits or withdrawals.
}}
\endgroup
\caption{A generated base profile example used in the rolling prediction.}
\label{fig:profile-example}
\end{figure}

\section{Examples in Human-Evaluation}
\label{app:explanation_alignment_cases}

The following cases illustrate strategy-alignment scores 1--5
and the distinction discussed in Finding~5.

\paragraph{Score 1.}
The explanation interprets stable portfolio exposure and repeated
price monitoring as a deliberate strategy to wait. These
observations do not establish such a rule, leaving the attributed
strategy unsupported.

\paragraph{Score 2.}
The explanation infers a recurring reversal strategy from an
isolated buy--sell sequence. The sequence provides limited support
for this interpretation because it does not establish a repeated
pattern.

\paragraph{Score 3: WMT/MSFT.}
The explanation captures selective position adjustment but
attributes the sales to profit-taking following recent
appreciation. The adjustments are observable; the stated motive
is not established by the records.

\paragraph{Score 4: BITO.}
Repeated additions support the explanation's description of
incremental accumulation. However, characterizing the holdings
as high-conviction goes beyond the available behavioral evidence.

\paragraph{Score 5: SKUU.}
Four same-session round trips support a recurring
repurchase-and-exit pattern, yielding strategy alignment of 5.
Decision-rationale alignment remains 4: the history establishes
the pattern but does not identify the condition for another
entry on the next trading day. This case illustrates how strong
strategy alignment can coexist with incomplete support for
a specific decision.

\section{Additional Experimental Content}
\label{app:evaluation_details}

Table~\ref{tab:appendix-class-metrics} reports detailed experimental results in Section~\ref{sec:rq1}.

\begin{table*}[t]
\centering
\footnotesize
\caption{
Detailed User Simulator Behavioral Metric Evaluation(\%). P and R denote precision and recall.
}
\label{tab:appendix-class-metrics}
\setlength{\tabcolsep}{4pt}

\begin{tabular*}{\textwidth}{
@{\extracolsep{\fill}}lrrrrrrr@{}
}
\toprule
Model
& \shortstack{Trade\\P}
& \shortstack{No-trade\\P}
& \shortstack{Trade\\R}
& \shortstack{No-trade\\R}
& \shortstack{Buy-only\\R}
& \shortstack{Sell-only\\R}
& \shortstack{Buy-and-sell\\R} \\
\midrule
Gemini-3-Flash
& 42.57 & 89.52 & 76.55 & 65.99
& 39.45 & 4.23 & 65.35 \\
Gemini-3.5-Flash
& 49.74 & 86.35 & 61.89 & 79.40
& 21.10 & 4.23 & 67.72 \\
GPT-4o-mini
& 51.58 & 85.73 & 58.63 & 81.87
& 30.28 & 7.04 & 51.97 \\
GPT-5.6-Luna
& 55.32 & 86.26 & 59.28 & 84.23
& 23.85 & 5.63 & 59.84 \\
GPT-5.6-Sol
& 64.85 & 81.38 & 34.85 & 93.78
& 11.01 & 8.45 & 41.73 \\
DeepSeek-V4-Flash
& 54.94 & 82.21 & 41.69 & 88.73
& 11.01 & 7.04 & 50.39 \\
Qwen3.8-27B
& 52.33 & 86.73 & 62.21 & 81.33
& 28.44 & 5.63 & 58.27 \\
\bottomrule
\end{tabular*}
\end{table*}

\subsection{Retrospective Portfolio Comparison}
\label{app:portfolio_comparison}

Observed and simulated portfolios share the same
initial holdings and cash.
Simulated transaction sizes use the median notional
of observed positive position increments and the
median proportional reduction in existing positions.
Calibration uses the available observation period,
making these comparisons retrospective.

Predicted sells precede buys and are limited to
available holdings.
Buy amounts are capped by an equal share of available
cash, without borrowing.
If calibration is unavailable, buys are omitted and
predicted sells liquidate the corresponding holdings.
Transactions use the latest available closing price
at or before the execution date; transactions without
prices are skipped.

Both trajectories use common adjusted prices and
compounded daily holding returns, with cash earning
zero return.
Unpriced positions are omitted from valuation.
The comparisons exclude deposits, withdrawals,
transaction costs, and intraday execution effects.
The displayed cases illustrate possible portfolio
divergence without estimating its prevalence
across participants.

\subsection{Observational Research--Trading Analysis}
\label{app:research_trading}

\paragraph{Opportunity set and exposure.}
The analysis unit is a participant--ticker--target-date
opportunity.
Eligible tickers are held or have been viewed or traded
during the preceding 30 calendar days, as recorded
at the trading-day baseline before exposure.
This opportunity set is fixed before the exposure day.
Research escalation requires quote attention within
the preceding seven trading days, no research in the
two baseline sessions, and research on the exposure day.
Research includes technical analysis, news, company
and earnings information, order-book inspection,
and options-related exploration.
The outcome is an executed same-ticker trade
on the next trading day.

\paragraph{Adjustment and estimation.}
We adjust for pre-exposure holdings, position returns,
information-seeking activity, trading history,
and available asset, market, and portfolio conditions.
Ridge logistic nuisance models include participant,
ticker, and target-date indicators, standardized
numerical covariates, and missing-information indicators.
An augmented inverse-probability-weighted estimator
produces the adjusted trading-probability difference.

Common support is the intersection of the exposed
and unexposed groups' 1st--99th propensity percentile
ranges, bounded by $[0.01,0.99]$.
This restriction retains 13,907 of 15,125 opportunities.
Stabilized weights are capped at the 99th percentile
for weighted contrasts and balance diagnostics.
Percentile 95\% intervals use 1,000 participant-clustered
bootstrap resamples with seed 23, conditioning on
the fitted nuisance models.

\paragraph{Identification diagnostics.}
We assess overlap, weighted covariate balance,
a pre-trend placebo, a wrong-ticker control,
and a future-research placebo.
The wrong-ticker control rotates outcomes within
participant--date opportunity sets.
The future-research placebo assesses temporal
asymmetry and is not a pure negative control.
Residual imbalance exceeds the 0.10 threshold
(maximum absolute weighted SMD: 0.324),
and the future-research placebo remains positive.
These diagnostics support reporting an adjusted
association without attributing subsequent trading
causally to research escalation.

\section{Rolling Prediction Example}
\label{app:prediction_prompt}

Figure~\ref{fig:prediction-example} illustrates a rolling next-day prediction instance including the point-in-time evidence, structured prediction, and natural-language explanation.

\begin{figure*}[t]
\centering
\begin{minipage}{0.94\textwidth}
\small

\noindent\textbf{A. Rolling Prediction Input}

\vspace{4pt}

\noindent
\begin{tabular}{@{}p{0.21\textwidth}p{0.70\textwidth}@{}}
\toprule
\textbf{Evidence} & \textbf{Observed Information} \\
\midrule

Prediction setup
&
Cutoff: Aug. 11, 2026 (market close);
Target: Aug. 12, 2026.
\\[3pt]

User profile
&
shown in Appendix~\ref{sec:user_profile}
\\[3pt]

Platform activity
&
333 interactions across 7 active days.
Most viewed: RKLX (36 views), NEBX (18 views).
\\[3pt]

Recent trading
&
Traded on all 7 recent sessions;
31 buy and 42 sell executions.
Both buying and selling occurred on all 7 sessions.
\\[3pt]

Portfolio holdings
&
BLSH, CRCL, DRAM, NVD, ORCL, SGOV,
SHV, SQQQ, TMDX, TSLL.
\\[3pt]

Portfolio changes
&
3 new positions, 6 additions,
4 trims, and 3 exits.
\\[3pt]

Market context
&
SPY: +1.70\%;
QQQ: +2.63\%;
portfolio-weighted return: +1.71\%.
\\[3pt]

Watchlist
&
NEBX: 18 views;
7-day return: $-20.80\%$.
\\

\bottomrule
\end{tabular}

\vspace{12pt}

\noindent\textbf{B. Rolling Prediction Output}

\vspace{4pt}

\noindent
\begin{tabular}{@{}p{0.21\textwidth}p{0.70\textwidth}@{}}
\toprule
\textbf{Prediction} & \textbf{Model Output} \\
\midrule

Trade activity
&
Trade
\\[3pt]

Trade side
&
Buy and sell
\\[3pt]

Ranked buy tickers
&
RKLX $\rightarrow$ PLTD $\rightarrow$ NEBX
\\[3pt]

Ranked sell tickers
&
RKLX $\rightarrow$ CRCL $\rightarrow$ TMDX
\\

\bottomrule
\end{tabular}

\vspace{12pt}

\noindent\textbf{C. Prediction Explanation}

\vspace{4pt}

\noindent
\begin{quote}
\small
\emph{
``A trade on both sides is likely because the user
consistently rotates short-horizon positions:
all seven recent sessions included buys and sells,
while the median holding period is 4.34 days.
RKLX is the strongest two-sided candidate,
with 36 views, ten recent orders, and repeated
buy-sell re-entry, including both sides on August 11.''
}

\medskip

\emph{
``Ticker-level timing remains uncertain because
no explicit exit rule or trade rationale is observed.''
}
\end{quote}

\end{minipage}

\caption{
Illustrative rolling next-day prediction example.
(A) Abbreviated prediction input.
(B) Structured prediction output.
(C) Natural-language explanation.
}
\label{fig:prediction-example}
\end{figure*}
\end{document}